\documentclass[runningheads]{llncs}
\usepackage{times}
\usepackage{microtype}
\usepackage{epsfig}
\usepackage{graphicx}
\usepackage{amsmath}
\usepackage{amssymb}
\usepackage{makecell}
\usepackage[]{algorithm2e}
\usepackage{algpseudocode}
\usepackage{booktabs}

\newcommand{\etal}{\textit{et al}.}

\begin{document}
\title{An Efficient Approximate kNN Graph Method for Diffusion on Image Retrieval}
\titlerunning{Efficient approx. kNN graph method for diffusion on image retrieval}
%
\author{Federico Magliani\inst{1}\orcidID{0000-0001-5526-0449} \and
Kevin McGuinness\inst{2} \and
Eva Mohedano \inst{2} \and
Andrea Prati\inst{1}}
\authorrunning{F. Magliani et al.}
%
\institute{IMP lab - University of Parma, Parma, Italy 
\email{federico.magliani@studenti.unipr.it}\\
\and
Insight Centre for Data Analytics - DCU, Dublin, Ireland.
}
\maketitle              

\begin{abstract}
The application of the diffusion in many computer vision and artificial intelligence projects has been shown to give excellent improvements in performance. One of the main bottlenecks of this technique is the quadratic growth of the kNN graph size due to the high-quantity of new connections between nodes in the graph, resulting in long computation times. Several strategies have been proposed to address this, but none are effective and efficient. Our novel technique, based on LSH projections, obtains the same performance as the exact kNN graph after diffusion, but in less time (approximately 18 times faster on a dataset of a hundred thousand images). The proposed method was validated and compared with other state-of-the-art on several public image datasets, including Oxford5k, Paris6k, and Oxford105k.
\end{abstract}

\section{Introduction}

Content-Based Image Retrieval (CBIR) is concerned with finding  the most similar images to a query in an image dataset, selected or photographed by the user. 
Recent improvements in features extraction through Convolutional Neural Networks (CNN) and algorithms for embedding, like the several R-MAC strategies \cite{tolias2015particular}, \cite{gordo2017end}, \cite{magliani2018accurate}, have made it possible to obtain excellent results on datasets of hundreds of thousand images in reasonable time~\cite{magliani2019landmark}.
Recently, the application of diffusion process on CBIR datasets have allowed boosting retrieval performance \cite{iscen2017efficient}: it permits finding more neighbour, that are close to the query on the nearest-neighbour manifold but not in  the Euclidean representation space. Diffusion propagates the similarities from a query point on a pairwise affinity matrix to all the dataset elements \cite{zhou2004ranking}.
To apply this process, it is necessary to create a kNN graph of all image embeddings in the dataset. Generally, the more discriminative the embeddings are, the better the results achievable through  diffusion.

\begin{figure}[t!]
\centering
\includegraphics[width= 0.75\textwidth]{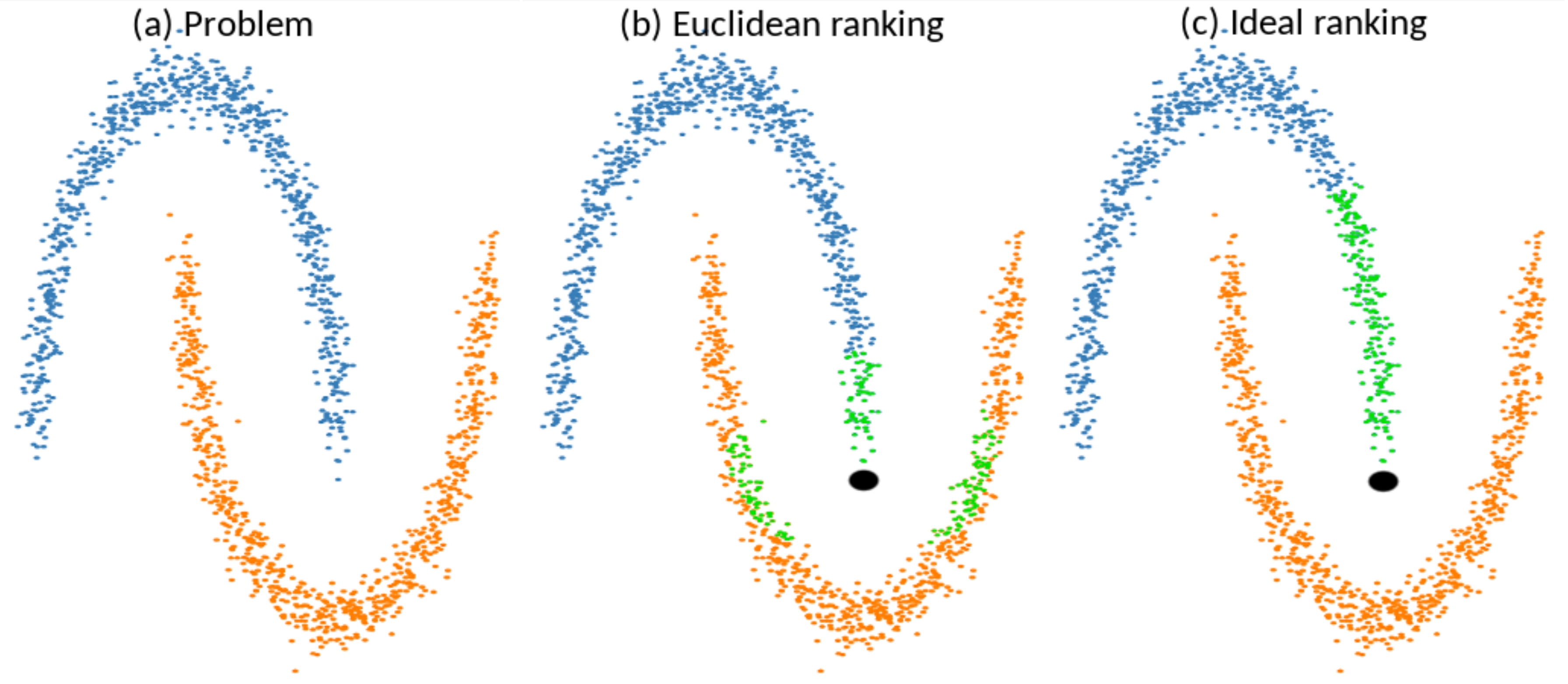}
\caption{Motivation of the use of diffusion approach for image retrieval tasks. In the first figure (a) two distributions of data are represented: in one distribution the points are blue-coloured and in the other the points are orange-coloured. In the second figure (b), the black point indicates the query point and the green points represent the results obtained after the execution of the retrieval exploiting the Euclidean distance. As you can see some correct results are retrieved (points belonging to the blue distribution), but also some incorrect results (points belonging to the orange distribution). Instead, the third figure (c) represents the ideal ranking for the indicated query point. This result can be easily obtained applying the diffusion process from the query point to the other elements of the dataset. Best viewed in color.}
\label{diffusion_example}
\end{figure}

Diffusion is an iterative process that simulates a random walk on the image similarity graph. It consists of walking on the graph, from the query point, with the objective of finding the best path, i.e. to retrieve the neighbours of the query point.
This is possible exploiting the weights of the edges of the kNN graph, which indicate the similarity between two nodes: the greater the weight, the more similar the two nodes are.

\par
Unfortunately, this recent method leads to new challenges: understanding the data distribution to correctly set the diffusion parameters, dealing with the size of the kNN graph, which grows quadratically with the dataset size, and reducing the convergence time for the resolution of the linear system related to the diffusion mechanism.

The kNN graph is needed to apply diffusion and the number of the edges in the graph is important for the final retrieval performance. Furthermore, it is impossible to know how many and which edges the graph needs for achieving good performance before applying diffusion. Therefore, a common strategy used in previous works was a brute-force approach, i.e. the graph is created with the connections between all the possible pairs of nodes.
Obviously, increasing the number of edges increases the size of the entire graph. For example, considering a dataset composed by $N$ images, the exact or brute-force graph will have $N \cdot N$ edges and the approach will have a complexity $O(N^2)$; if $N = 100K$, the number of edges will be 10 billion. \par
Several methods proposed to implement an approximated method for the creation of the kNN graph, drastically reducing the computational time \cite{sieranoja2018fast}, \cite{dong2011efficient}, \cite{chen2009fast}, \cite{zhang2013fast}.
\par
Following this idea that not all the edges are necessary for creating the kNN graph, we propose a fast approach for the creation of an approximate version of the kNN graph, based on LSH projections \cite{indyk1998approximate}, which maintains only the useful edges to reduce computation time and memory requirements. The new graph achieves the same retrieval results as the exact kNN graph after diffusion \cite{zhou2004ranking} on several public image datasets, but does so more efficiently. 

The main contributions of this paper are:

\begin{itemize}
\item An efficient algorithm based on LSH projections for the creation of an approximate kNN graph that obtains the same performance as brute-force in less time.
\item Several optimizations in the implementation that reduce the computational time and memory use.
\item The use of the multi-probe LSH \cite{lv2007multi} for improving diffusion performance on the kNN graph.
\end{itemize}


\section{Related work} \label{ref:related}
Graphs have been used for different tasks in computer vision: applying diffusion for image retrieval \cite{iscen2017efficient},  unsupervised fine-tuning \cite{iscen2018mining}, propagating labels for semi-supervised learning \cite{douze2018low}, creating manifold embeddings \cite{xu2017iterative}, and training classifiers exploiting the cycles found in the graph \cite{li2016unsupervised}. 

In particular, a k-Nearest Neighbor (kNN) graph is an undirected graph $G$ denoted by $G(V,E)$, where $V$ represents the set of nodes $V=\left\{(v_1, v_2, \dots , v_n)\right\}$ and $E$ represents the set of edges $E=\left\{(e_1, e_2, \dots , e_n)\right\}$. 
The nodes represent all the images in the dataset and the edges represent the connections between nodes. The weight of each edge determines how much the two images are similar: the higher the weight, the more similar the two images are. The weights of the edges are set using cosine similarities between embeddings.

The problem of kNN graph creation has been addressed in the literature in several ways. The simplest and naive approach is  brute-force, which tends to be very slow but usually obtains the best results. 

To speed up the process while retaining good retrieval accuracy, approximated kNN graphs have been used. The methods for the construction of the approximate kNN graph can be divided in two strategies: methods following the strategy of divide and conquer, and methods using local search  (e.g., NN-descent \cite{dong2011efficient}). Divide-and-conquer methods generally consist of two stages: subdividing the dataset in parts (divide), and creating a kNN graph for each sample followed by merging all subgraphs to form the final kNN graph (conquer).

\par
As foreseeable, the performance and the computational time depend on the number of subdivisions. 
The most famous approach following the divide and conquer is based on Locality Sensitive Hashing (LSH) projections \cite{indyk1998approximate} for creating the approximate kNN graph \cite{zhang2013fast}. The authors in \cite{zhang2013fast} used a spectral decomposition of a low-rank graph matrix that needs much time because it is supervised. Chen \etal \cite{chen2009fast} follow the same strategy, but they apply recursive Lanczos bisection \cite{chen2009fast}. They proposed two divide steps: the overlap and the glue method. In the former, the current set is divided in two overlapping subsets, while in the latter, the current set is divided into two disjoint subsets and a third, called gluing set, is used to merge the two resulting disjoint subsets.
Wang \etal \cite{wang2012scalable} implemented an algorithm for the creation of an approximate kNN graph through the application in several iterations of the divide-and-conquer strategy. The peculiarity of the method is that the subsets of the dataset elements used during the divide phase are randomly chosen. Repeating several times this process allows to theoretically cover the entire dataset. Our approach differs from these approaches in the way the LSH projections are created. Moreover, the strategy followed for creating the graph is different and more efficient.

\par
Regarding the second strategy (local search), Dong \etal \cite{dong2011efficient} proposed an approach called NN-descent \cite{dong2011efficient}, based on the idea that ``a neighbour of my neighbour is my neighbour.'' For each image descriptor, a random kNN list is created. The algorithm starts searching random pairs on the kNN list, then it calculates the similarity between elements and finally updates the kNN list of these elements. This process continues until the number of updates is smaller than a threshold.
Obviously, by increasing the number of neighbours contained in the kNN list it tends to the brute-force approach; therefore, the trade-off between the speed and accuracy performance needs to be correctly evaluated.
Park \etal \cite{park2013scalable},  Houle \etal \cite{houle2014improving} and Debatty \etal \cite{debatty2014building} proposed variations to the NN-descent, by adapting the basic approach to their specific application domains.
Sieranoja \etal \cite{sieranoja2018fast} proposed a solution, called Random Pair Division, that exploits both the divide and conquer and the NN-descent techniques. The division of the dataset in subsamples is executed through the random selection of two dataset descriptors: if the descriptor to be clusterized is close to the first one, it will be put in the first set, otherwise in the second one. After that, all image descriptors are clustered: if the size of each set is greater than a threshold, the subdivision process continues in the same way only for this large set. The conquer phase is executed through the application of the brute-force approach. In the end, a one-step neighbour propagation is applied for improving the final performance. It also exploits the principle of NN-descent: similar nodes that are not connected will be connected.

\section{Proposed approach: LSH kNN graph} \label{ref:proposed_approach}

The proposed approach uses LSH projections to divide the global descriptors of the dataset in many subsets. We first explain LSH and then detail a fast and efficient solution for kNN graph creation for diffusion in image retrieval.

\subsection{Notations and background of LSH}

Locality-Sensitive Hashing (LSH) \cite{indyk1998approximate} is one of the first hashing technique proposed for compression and indexing tasks. After the creation of some projection functions, it allows projection of points close to each other into the same bucket with high probability. It is defined as follows \cite{wang2014hashing}: a family of hash functions $\mathcal{H}$ is called $(R,cR,P_{1},P_{2})$-sensitive if, for any two items $\mathbf{p}$ and $\mathbf{q}$, it holds that: 
\begin{itemize}
\item if $\mathrm{dist}(\mathbf{p},\mathbf{q})\leq R,$ $\mathrm{Prob}[h(\mathbf{p})=h(\mathbf{q})]\geq P_{1}$ 
\item if $\mathrm{dist}(\mathbf{p},\mathbf{q})\geq cR,$ $\mathrm{Prob}[h(\mathbf{p})=h(\mathbf{q})]\leq P_{2}$ 
\end{itemize}
\noindent with $c>1$, $P_{1}>P_{2}$, $R$ is a distance threshold, and $h(\cdot)$ is the hash function. In other words, the hash function $h$ must satisfy the property to project ``similar" items (with a distance lower than the threshold R) to the same bucket with a probability higher than $P_{1}$, and have a low probability (lower than $P_{2}<P_{1}$) to do the same for ``dissimilar" items (with distance higher than $cR>R$).

The hash function used in LSH for Hamming space embedding is a scalar projection:
$$
h(\mathbf{x})=\mathrm{sign}(\mathbf{x}\cdot\mathbf{v})
$$
where $\mathbf{x}$ is the feature vector and $\mathbf{v}$ is a fixed random vector sampled from an $D$-dimensional isotropic Gaussian distribution $\mathcal{N}(0,I)$.
The hashing process is repeated $L$ times, with different Gaussian samples to increase the probability of satisfying the above constraints. 


Subsequently, different hashing techniques may be implemented. The multi-probe LSH \cite{lv2007multi} has the objective 
to reduce the number of hash tables used for the projections, exploiting the fundamental principle of LSH that similar items will be projected in the same buckets or in near buckets with high probability. During the search phase, multi-probe LSH checks also the buckets near the query bucket. In the end, this approach allows to improve the final performance, but it increases the computational time.

\subsection{Basic algorithm for kNN graph construction}

Given a dataset $\mathcal{S} = \{s_1, \dots , s_N\}$, composed by $N$ images, and a similarity measure $\theta : \mathcal{S} \times \mathcal{S} \rightarrow \mathbb{R}$, the kNN graph for $\mathcal{S}$ is a undirected graph $G$, that contains edges between the nodes $i$ and $j$ with the value from the similarity measure $\theta(s_i, s_j) = \theta(s_j, s_i)$. The similarity measure can be calculated in different ways related to the topic. In this case we use the cosine similarity as a metric, so the similarity is calculated through the application of the dot product between the image descriptors.


Our approach, called LSH kNN graph, follows the idea to first subdivide the entire dataset in many subsets, based on the concept of similarity between the images contained in the dataset. This process is done through the use of LSH projections and allows creation of a set of buckets $B = \{B_1, \dots, B_N\}$ from several hash tables $L = \{L_1, \dots , L_M\}$. The number of buckets $N$ depends on the bits used for the projection ($\delta$) and to the number of the hash tables ($M$): $N = 2^\delta \cdot M$. Each of these buckets will contain the projected elements $B_1 = \{b_{11}, \dots , b_{1n}\}$. The result of this process represents an approximate result because it is not generally possible to project every element the dataset into the same bucket an as all its neighbours. It is therefore necessary to find a trade-off between the number of the buckets for each hash table ($2^\delta$), modifying the bits used ($\delta$) for the hashing, and the number of hash tables ($M$) used for the projection. 
For this first task, using a small number of buckets allows to project more data in the same bucket. It reduces the computational time for this phase, but it increases the entire computation of the overall approach. On the other hand, a high number of buckets in each hash table increases the computational time of this step, but it reduces overall computation. 
\par
For each bucket containing a subset of the dataset, a brute-force graph with edges $G_i = \{  (b_{ix},b_{iy}, \theta(b_{ix},b_{iy})) : (b_{ix},b_{iy}) \in B_i  \}$ is constructed. Applying a brute-force construction on many subsets is faster than apply one time the brute-force on the entire dataset.
In the end all the subgraphs need to be merged in the final graph $G = G_1 \cup \dots \cup G_N$.
\par
Unlike the usual LSH \cite{zhang2013fast} based method, the proposed approach does not follow exactly the divide and conquer strategy. 
LSH projections are applied for dividing the dataset into subsets, but for reducing the computational time it is preferable to start creating the final graph, instead to create many approximate kNN graph and then merge them using the one-step neighbour propagation algorithm. 
The number of elements to sort in the kNN list and the number of similarity scores to calculate improves the quality of the final graph and the retrieval accuracy, but also reduces the computational time.

\subsection{Multi-probe LSH}

We also propose a multi-probe version, called multi LSH kNN graph, to reduce the number of hash tables used. 
Unlike the classic multi-probe LSH  algorithm \cite{lv2007multi}, in which the system checks  neighboring buckets during the search phase, here all the elements are projected in the neighbors buckets during the projection phase, but only in the 1-neighbourhood. This represents the set of buckets differing by one bit  to the analysed bucket (i.e. with Hamming distance $H_d \le 1$).
More formally, the elements obtained with the application of the multi-probe LSH are the followings: $$B_{\text{multi-probe}} = \{b_{x1}, \dots , b_{xp}  : H_d(b_{\text{query}},b_{xj}) \le 1 ,\, b_{xj} \in B,\, 0 \le x \leq P \}$$

Note that the number of neighbours of each bucket scales with the bits used for the projection as: $\sum_{i=0}^{l}{{\log_{2}\delta} \choose {i}}$. Even though it increases the final retrieval performance, this approach requires more time for the kNN graph creation than the previous one. To get a good trade-off between the computational time required for the similarity measure calculations and the quality of the final graph, only a percentage $\gamma$ of the elements projected on the 1-neighbour buckets are retained. The best trade-off is reached using $\gamma = 50\%$, which means that each element is also projected randomly in the half of its 1-neighbour buckets.

During the \textit{conquer} phase, as in the previous LSH kNN graph method, all the pairs of the indexes of the images found in the buckets will be connected through the calculation of the similarity measure.

\section{Experimental results} \label{ref:results}

Previous works have evaluated the methods for creating approximate kNN graphs by checking the number of common edges between the approximate and the exact kNN graph.	
In our case the kNN graph pipelines are evaluated after the diffusion and retrieval modules. 
The diffusion approach adopted and the parameters configuration is the same of the work of Iscen et al. \cite{iscen2017efficient}.
The rationale for this lies in our objective to evaluate how effective (and efficient) are our proposals for the approximate kNN graph creation in terms of retrieval accuracy when diffusion is applied. 
 R-MAC descriptors \cite{iscen2017efficient} are as image embeddings in all the experiments for the creation of the kNN graphs.

\subsection{Datasets}
There are many different image datasets for Content-Based Image Retrieval that are used to evaluate  algorithms. The most used are:
\begin{itemize}
\item \textbf{Oxford5k} \cite{Philbin07} containing 5063 images, subdivided in 11 classes. All the images are used as database images and there are 55 query images, which are cropped to make the querying phase more difficult;
\item \textbf{Paris6k} \cite{Philbin08} containing 6412 images, subdivided in 12 classes. All the images are used as database images and there are 55 query images, again cropped;
\item \textbf{Oxford105k} that is a combination of Oxford5k and 100k distractors from Flickr1M\cite{huiskes2008mir}. 
\end{itemize}

\subsection{Evaluation metrics}
Mean Average Precision (mAP) is used on all datasets to evaluate the retrieval accuracy. mAP identifies how many elements retrieved at different score thresholds are relevant to the query image, and averages this across all queries. We use $L_2$ distances to compare query images with the database.

\subsection{Sparse matrices for kNN graph}

It is worth emphasizing that there are a lot of null values in the affinity matrix. In fact, on Oxford5k the approximate kNN graph constructed with LSH kNN graph method has only the $0.7\%$ of the edges of the brute-force graph.
Furthermore, not all the similarity measure are useful for the diffusion process, suggesting to remove or avoid to insert edges with weight less than a threshold ($th$), without jeopardizing the final retrieval performance. Hence, each element of the matrix can be represented as following:

$$g_{ij} =
\bigg \{
\begin{array}{ll}
\theta(s_i, s_j) & \textrm{   if  } \theta(s_i, s_j) \ge th \\
0 & \textrm{   otherwise} \\
\end{array}
$$

From our experiments, this threshold can be set to $0.3$.
Given the high number of null values in the affinity matrix, sparse matrices can be used to reduce the computational time and still obtain good results also on large datasets.
Moreover, considering that the matrix is symmetric, only the upper or lower values of the matrix are needed:

$$g_{ij} =
\bigg \{
\begin{array}{ll}
\theta(s_i, s_j) & \textrm{   if   } j \ge i \land \theta(s_i, s_j) \ge th \\
0 & \textrm{   otherwise  }  \\
\end{array}
$$

If the similarity value is missing, the row and the column are switched.

\subsection{Implementation details}
Two different types of sparse matrix has been tested: Compressed Row Storage (CRS) format and Coordinate (COO) format \cite{golub2012matrix}.
The CRS sparse matrix is composed by three vectors: values (containing the values of the dense matrix different from zero); column indexes (containing the column indexes of the elements contained in the values vector); and row pointers (containing the locations of the values vector that indicate the beginning of a new row). Instead, the COO sparse matrix is composed by three vectors: a vector representing the non-zero elements (the values), the row and the column coordinate of each value contained in the values vector.
The second solution is simpler than the first to implement, but it requires more space on disk.

However, using hash tables, it happens that the same edge weight is inserted multiple times. Therefore, every time a new value is inserted in a CRS matrix, checking whether the value is already in the matrix might be a possible solution. Unfortunately, this tends to be a time consuming process. Conversely, using a COO matrix, all the values (including repeated ones) are inserted, but a sorting is performed and duplicates are removed. Applying once the sorting and removing the duplicates is faster than performing $N \cdot L$ times the search, given that sorting has a $O(N\log_2N)$ complexity which is lower than the $O(N)$ complexity of the search.


\subsection{Results on Oxford5k}
Table \ref{results_oxf5k} reports the retrieval results after diffusion application of different kNN graph techniques. Note that changing the values of LSH ($\delta$ and $L$) produces different results. The best configuration is $\delta = 6$ and $L = 2$ applying the multi LSH kNN graph approach. the best trade-off between the computational time for kNN graph creation and the final retrieval performance is the LSH kNN graph with  $\delta = 6$ and $L = 20$. NN-descent produces good results, but it needs a lot of time for the graph creation (55 seconds). Furthermore, it does not obtain results comparable to the brute-force or the LSH kNN graph method.

\begin{table*}[htb]
\centering
\setlength{\tabcolsep}{6pt}
    \begin{tabular}{lccc}
    \toprule \textbf{Method} &  \textbf{LSH projection} & \textbf{kNN graph creation} & \textbf{mAP}  \\
    \midrule
	LSH kNN graph ($\delta=6, L=20$)  & 0.45 s & \textbf{0.52 s} & 90.95\% \\ 
	LSH kNN graph ($\delta=8, L=10$)   & 0.4 s & 0.95 s & 88.98\% \\ 
	multi LSH kNN graph ($\delta=6, L=2$)   & 0.29 s & 1.54 s & \textbf{91.13\%} \\ 
	NN-descent \cite{dong2011efficient}* &  - & 55 s & 83.81\% \\ 
	RP-div \cite{sieranoja2018fast} (size = 50)*  & - & 1.16 s & 82.68\% \\ 
	Wang \etal \cite{wang2012scalable}* & - & 1.5 s & 90.60\% \\ 
	brute-force  & - & 1.33 s & 90.79\% \\ 
	\bottomrule
   \end{tabular}
        \caption{Comparison of different approaches of kNN graph creation tested on Oxford5k. * indicates that the method is a C++ re-implementation.}

        \label{results_oxf5k}
\end{table*}

RP-div \cite{sieranoja2018fast} is very fast but collecting random elements from the dataset for the divide task does not give good results in retrieval after diffusion.

The method implemented by Wang \etal \cite{wang2012scalable} obtains a different result each execution, so the reported performance is the average of ten experiments. The approach is very fast, but did not achieve the best mAP. Note also that the brute-force method is executed on GPU, instead all the other methods are executed on CPU.

We also perform some experiments with regional descriptors (Table \ref{results_oxf5k_regional}). The use of regional descriptors demonstrates an improvement on the final performance due to high number of descriptors for each image (usually 21). In this case the total number of descriptors used for the creation of the kNN graph are approximately 100K. Note we omit testing RP-div and NN-descent here due to poor previous accuracy/computation performance.


\begin{table*}[t]
\centering
\setlength{\tabcolsep}{6pt}
    \begin{tabular}{lccc}
    \toprule
    \textbf{Method} &    \textbf{LSH projection} & \textbf{kNN graph creation} & \textbf{mAP}  \\ \midrule
	LSH kNN graph ($\delta$=6, L=20) & 9 s & 100 s & \textbf{94.67\%} \\ 
	LSH kNN graph ($\delta$=8, L=10) & 6 s & \textbf{45 s} & 93.68\% \\ 
	multi LSH kNN graph ($\delta$=6, L=2)  & 6 s & 350 s & 93.96\% \\ 
	Wang \etal \cite{wang2012scalable}* & - & 148 s & 91.69\% \\ 
	brute-force & - & 15816 s & 93.80\% \\ 
	\bottomrule
   \end{tabular}
        \caption{Comparison of different approaches of kNN graph creation tested on Oxford5k using regional R-MAC descriptors. * indicates that the method is a C++ re-implementation.}
\label{results_oxf5k_regional}

\end{table*}

\subsection{Results on Paris6k}
Table \ref{results_par6k} shows the results on the Paris6k dataset, which are similar to those obtained to Oxford5k. 
\begin{table*}[ht]
\centering
\setlength{\tabcolsep}{4pt}
    \begin{tabular}{lccc }
    \toprule
    \textbf{Method} &   \textbf{LSH projection} & \textbf{kNN graph creation} & \textbf{mAP}  \\ \midrule
	LSH kNN graph ($\delta=6, L=20$)  & 1 s & 0.80 s & \textbf{97.01\%} \\ 
	LSH kNN graph ($\delta=8, L=10$) &  0.78 s & \textbf{0.28 s} & 95.93\% \\ 
	multi LSH kNN graph ($\delta=6, L=2$) & 0.35 s & 2.28 s & 96.81\% \\ 
	NN-descent \cite{dong2011efficient} (neighbours = 50)*  & - & 60.10 s & 94.24\% \\ 
	RP-div \cite{sieranoja2018fast} (size = 50)*  & - & 3.63 s & 96.25\% \\ 
	Wang \etal \cite{wang2012scalable}* & - & 1.95 s & 96.75\% \\ 
	brute-force & - & 1.81 s & 96.83\% \\ 
	\bottomrule
   \end{tabular}
        \caption{Comparison of different approaches of kNN graph creation tested on Paris6k. \hspace{20mm} * indicates that the method is a C++ re-implementation.}
        \label{results_par6k}
\end{table*}

However, in this case, LSH kNN graph is the fastest approach and also it obtains the best retrieval performance after the application of diffusion. Multi-LSH kNN method obtains a good result, but in more time than the brute-force approach. 

\subsection{Results on Oxford105k}
Table \ref{results_oxf105k} reports results for the experiments executed on Oxford105k. again RP-div and NN-descent are not tested due to poor trade-off previously obtained.
\begin{table*}[ht]
\centering
\setlength{\tabcolsep}{6pt}
    \begin{tabular}{lccc}
    \toprule
    \textbf{Method} &    \textbf{LSH projection} & \textbf{kNN graph creation} & \textbf{mAP}  \\ \midrule
	LSH kNN graph ($\delta=6, L=20$) & 23 s & \textbf{77 s} & 92.50\% \\ 
	LSH kNN graph ($\delta=8, L=10$)  & 15 s & 145 s & 90.79\% \\ 
	multi LSH kNN graph ($\delta=6, L=4$) & 5 s & 420 s & \textbf{92.85\%} \\ 
	Wang \etal \cite{wang2012scalable}* & - & 150 s & 91.00\% \\ 
	brute-force  & - & 4733 s & 91.45\% \\ 
	\bottomrule
   \end{tabular}
        \caption{Comparison of different approaches of kNN graph creation tested on Oxford105k.  \hspace{20mm}            * indicates that the method is a C++ re-implementation.}
        \label{results_oxf105k}
\end{table*}
Increasing the dimension of the dataset illustrates the difference in  accuracy and computational time between the proposed approach and brute-force.
The proposed approaches obtain better results and trade-offs than other methods. In particular, LSH kNN graph ($\delta = 6$ and $L = 20$) achieves 92.50\% in only 77s for the graph creation process. The multi LSH kNN graph needs more time than the previous approach, but it reaches the best mAP on this dataset equals of 92.85\%.



\section{Conclusions}

We presented an algorithm called LSH kNN graph for the creation of an approximate kNN graph exploiting LSH projections. First, the elements of the dataset are subdivided in several subsets using an unsupervised hashing function and, then, for each one of the subsets a subgraph is created applying the brute-force approach.
The application of this algorithm with sparse matrices achieves very good results even on datasets that with a large number of images.
The proposed methods can generate a kNN graph faster than the brute-force approach and other state-of-the-art approaches, obtaining the same or better accuracy results after diffusion.
Furthermore, another version of the algorithm called multi LSH kNN graph was proposed, which uses multi-probe LSH instead of LSH for the subdivision of the elements in the subsets, increasing the quality of the final graph due to the greater number of elements found in the buckets of the hash tables.
In future work, we are pursuing the distribution these approaches across several machines to allow processing even larger datasets.

\section*{Acknowledgment}
This is work is partially funded by Regione Emilia Romagna under the ``Piano triennale alte competenze per la ricerca, il trasferimento tecnologico e l'imprenditorialit\`a".

{\small
\bibliographystyle{splncs04}
\bibliography{egbib}
}

\end{document}